\documentclass{llncs}

\usepackage{graphicx}
\usepackage{latexsym}
\usepackage{soul}
\usepackage{comment}
\usepackage{graphicx}
\usepackage{enumitem}
\usepackage{hyperref}
\usepackage{color}
\usepackage[colorinlistoftodos]{todonotes}
\usepackage{booktabs} 
\usepackage{multirow}

\usepackage{amssymb}

\newcommand{\ct}{\texttt{CheckThat! }}

\definecolor{darkturquoise}{rgb}{0.0, 0.81, 0.82}
\definecolor{lightblue}{rgb}{.50,.95,1}
\definecolor{tri}{rgb}{.25,.88,.82}
\definecolor{lilac}{rgb}{0.85,0.64,0.85}

\title{\ct at CLEF 2020:\\ Enabling the Automatic Identification\\ and Verification of Claims in Social Media}

\author{%
Alberto Barr\'on-Cede\~no\inst{1}\and
Tamer Elsayed\inst{2} \and
Preslav Nakov\inst{3} \and
Giovanni Da San Martino\inst{3}\and
Maram Hasanain\inst{2}\and
Reem Suwaileh\inst{2}\and
Fatima Haouari\inst{2}
}
\authorrunning{A. Barr\'on-Cede\~no et al.}
\institute{%
DIT--Universit\`a di Bologna, Forl\`i, Italy \and 
Qatar University, Doha, Qatar \and
Qatar Computing Research Institute, HBKU, Doha, Qatar   \\
\email{a.barron@unibo.it}   \\
\email{\{telsayed,maram.hasanain,rs081123,200159617\}@qu.edu.qa}\\
\email{\{pnakov,gmartino\}@hbku.edu.qa}
}
\begin{document}

\maketitle
\begin{abstract}
We describe the third edition of the \ct Lab, which is part of the 2020 Cross-Language Evaluation Forum (CLEF). \ct proposes four complementary tasks and a related task from previous lab editions, offered in English, Arabic, and Spanish. 
Task 1 asks to predict which tweets in a Twitter stream are worth fact-checking.  
Task 2 asks to determine whether a claim posted in a tweet can be verified using a set of previously fact-checked claims. 
Task 3 asks to retrieve text snippets from a given set of Web pages that would be useful for verifying a target tweet's claim. 
Task 4 asks to predict the veracity of a target tweet's claim using a set of Web pages and potentially useful snippets in them.
Finally, the lab offers a fifth task that asks to predict the check-worthiness of the claims made in English political debates and speeches. \ct features a full evaluation framework.  The evaluation is carried out using mean average precision or precision at rank $k$ for ranking tasks, and F$_1$ for classification tasks.
\end{abstract}

\section{Introduction} 
\label{sec:overview}

The mission of the \ct lab is to foster the development of technology that would enable the automatic verification of claims. 
Automated systems for claim identification and verification can be very useful as supportive technology for investigative journalism, as they could provide help and guidance, thus saving time~\cite{RANLP2017:debates,Hassan:15,hassan2017claimbuster,RANLP2019:checkworthiness:multitask}.
A system could automatically identify check-worthy claims, make sure they have not been fact-checked already by a reputable fact-checking organization, and then present them to a journalist for further analysis in a ranked list. 
Additionally, the system could identify documents that are potentially \textit{useful} for humans to perform manual fact-checking of a claim, and it could also estimate a \emph{veracity score} supported by evidence to increase the journalist's understanding and the trust in the system's decision. 

\ct at CLEF 2020 is the third edition of the lab.
The 2018 edition~\cite{clef2018checkthat} of \ct focused on the identification and verification of claims in political debates.\footnote{\url{http://alt.qcri.org/clef2018-factcheck/}}
Whereas the 2019 edition~\cite{ecir-checkthat:2019,clef-checkthat:2019}
also focused on political debates, isolated claims were considered as well, in conjunction with a closed set of Web documents to retrieve evidence from.\footnote{\url{https://sites.google.com/view/clef2019-checkthat}}

In 2020, \ct turns its attention to social media ---in particular to \emph{Twitter}--- as information posted on that platform is not checked by an authoritative entity before publication and such information tends to disseminate very quickly. Moreover, social media posts lack context due to their short length and conversational nature; thus, identifying a claim's context is sometimes key for enabling effective fact-checking~\cite{cazalens2018content}.

\section{Description of the Tasks}
\label{sec:tasks}
\begin{figure}[t]
\centering
\includegraphics[width=\columnwidth]{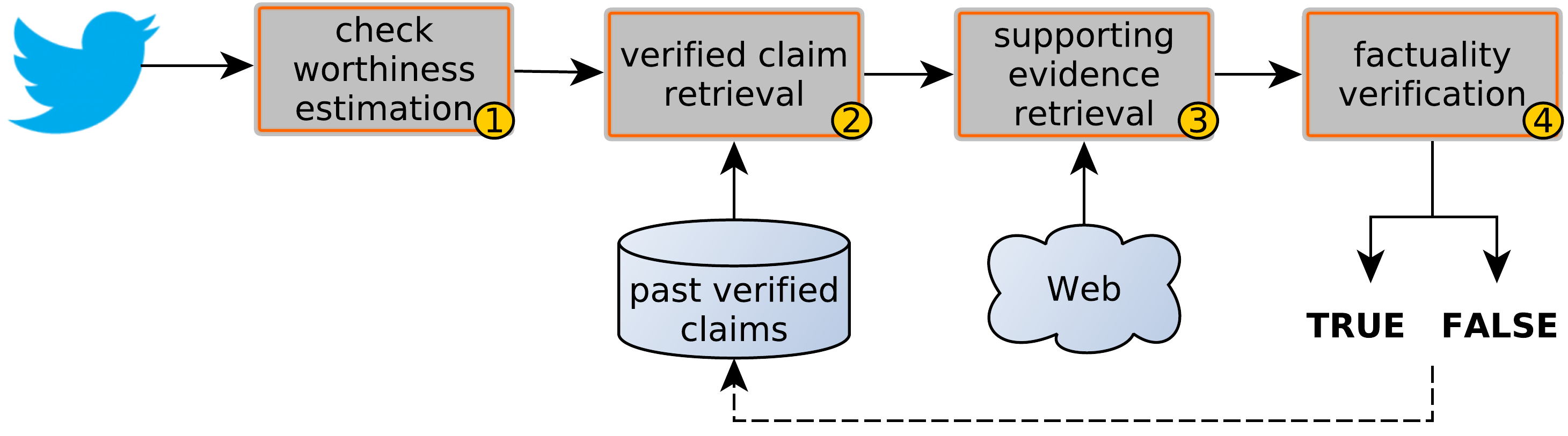}
\caption{Information verification pipeline. Our tasks cover all four steps. (Box 1 maps to task 1 whereas boxes 3--4 map to task 2 of the 2018 and 2019 editions~\cite{clef-checkthat:2019,clef2018checkthat}.)}
\label{fig:pipeline}
\end{figure}

\vspace{-0.25cm}
The lab is mainly organized around four tasks, which correspond to the four main blocks in the verification pipeline, as illustrated in Figure~\ref{fig:pipeline}. Tasks 1, 3, and 4 can be seen as reformulations of corresponding tasks in 2019, which enables re-use of training data and systems from previous editions of the lab (cf.\ Section~\ref{sec:prevct}).
Task~2 runs for the first time. 
While Tasks~1--4 are focused on Twitter, Task~5 (not in Figure~\ref{fig:pipeline}) focuses on political debates as in the previous two editions of the lab.  All tasks are run in English. Additionally, Tasks 1, 3, and 4 are also offered in Arabic and/or Spanish.

\vspace{-0.25cm}
\subsection{Task 1: Check-Worthiness on Tweets}
\label{sub:task1}

Task 1 is formulated as follows:
\textit{Given a topic and a stream of potentially-related tweets, rank the tweets according to their check-worthiness for the topic.} 

Previous work on check-worthiness focused primarily on political debates and speeches, but here we focus on tweets instead. 

\vspace{-0.25cm}
\subsubsection{Dataset}
We include ``topics'' this year, as we want to have a scenario that is close to that from 2019: a topic gives a context just like a debate did. 
%
We construct the dataset by tracking a set of manually created topics in Twitter. A sample of tweets from the tracked stream (per topic) is shared with the participating systems as input for Task 1. The systems are asked to submit a ranked list of the tweets for each topic. 
%
%
Finally, using pooling, a set of tweets is selected and then judged by in-house annotators. 

    

\vspace{-0.25cm}
\subsubsection{Evaluation}

We treat Task 1 as a ranking problem. Systems are evaluated using ranking evaluation measures, namely Mean Average Precision (MAP) and precision at rank $k$ (P@$k$). The official measure is P@$30$.

\vspace{-0.25cm}
\subsection{Task 2: Verified Claim Retrieval} 
Task 2 is defined as follows: \textit{Given a check-worthy claim and a dataset of verified claims, rank the verified claims, so that those that verify the input claim (or a sub-claim in it) are ranked on top.}


Given an input claim $c$ and a set $V_c=\{v_i\}$ of verified claims, we consider each pair $(c,v_i)$ as \textit{Relevant} if $v_i$ would save the process of verifying $c$ from scratch, 
and as \textit{Irrelevant} otherwise. 
Note that there might be more than one \textit{Relevant} verified claim per input claim, e.g.,~because the input claim might be composed of multiple claims. 
The task is similar to paraphrasing and textual similarity tasks, as well as to textual entailment~\cite{cer-etal-2017-semeval,Filice2015,nakov-EtAl:2016:SemEval}. 
\vspace{-0.25cm}
\subsubsection{Dataset}

Verified claims are retrieved from fact-checking websites such as \textit{Snopes} and \textit{PolitiFact}.
\vspace{-0.25cm}
\subsubsection{Evaluation}

Mean Average Precision on the first 5 retrieved claims (MAP@5) is used to assess the quality of the rankings submitted by the participants. 
A perfect ranking will have on top all $v_i$ such that $(c,v_i)$ is \textit{Relevant}, in any order, followed by all \textit{Irrelevant} claims. 
In addition to MAP@$5$, we also report MRR, MAP@$k$ ($k=3,10,20,all$) and Recall@$k$ for $k=3,5,10,20$ in order to provide participants with more information about their systems. 

\vspace{-0.25cm}
\subsection{Task~3: Evidence Retrieval}
\label{sub:task3}
Task~3 is defined as follows: \textit{Given a check-worthy claim on a specific topic and a set of text snippets extracted from potentially-relevant webpages, return a ranked list of all evidence snippets for the claim. Evidence snippets are those snippets useful to judge the claim's factuality}.

\vspace{-0.25cm}
\subsubsection{Dataset}
While tracking on-topic tweets, we search the Web to retrieve top-$m$ Web pages using topic-related queries. 
This would ensure the freshness of the retrieved pages and enable reusability of the dataset for real-time verification tasks.
%
Once we acquire annotations for Task 1, we share with  participants the Web pages and text snippets from them solely for the check-worthy claims, which would enable the start of the evaluation cycle for Task 3. 
In-house annotators will label each snippet as evidence or not for a target claim.
\vspace{-0.25cm}
\subsubsection{Evaluation}
Tasks~3 is a ranking problem. We evaluate the ranked list per topic using MAP and P@$k$. The official measure is P@10.

\vspace{-0.25cm}
\subsection{Task 4: Claim Verification} 
\label{sub:task4}
Task~4 is defined as follows: \textit{Given a check-worthy claim from a tweet and a set of potentially relevant Web pages, predict the veracity of the claim.} 

This task closes the verification pipeline.

\subsubsection{Dataset}
The dataset for this task is the same as for Task 3. The only difference is that the in-house annotators judge each claim as true or false.
\vspace{-0.25cm}
\subsubsection{Evaluation}
Task~4 is a binary classification problem. Therefore, it is evaluated using standard classification evaluation measures: Precision, Recall, $F_1$, and Accuracy. The official measure is macro-averaged $F_1$.

\vspace{-0.25cm}
\subsection{Task 5: Check-Worthiness on Debates} 

Task~5 is defined as follows: \textit{Given a debate segmented into sentences, together with speaker information, prioritize sentences for fact-checking.}

This is a ranking task and each sentence should be associated with a score. 
\vspace{-0.25cm}
\subsubsection{Dataset}
This is the third iteration of this task. We believe it is important to keep it alive as we have a large body of annotated data already and new material arrives with the coming 2020 US Presidential elections.

\vspace{-0.25cm}
\subsubsection{Evaluation}

Task~5 is yet another ranking problem. 
We use MAP as the official evaluation measure. We further report P@$k$ for $k \in \{5, 10, 20, 50\}$.

\section{Previously on \ct}
\label{sec:prevct}

\vspace{-0.25cm}
Two editions of \ct have been held so far. While the datasets come from different genres, some of the tasks in the 2020 edition are reformulated. Hence, considering some of the most successful approaches applied in the past represents a good starting point to address the current challenges.

\subsection{\ct 2019}

The 2019 edition featured two tasks~\cite{clef-checkthat:2019}:
\vspace{-0.25cm}
\paragraph{Task~1$_{2019}$.} \emph{Given a political debate, interview, or speech, transcribed and segmented into sentences, rank the sentences by the priority with which they should be fact-checked.}

The most successful approaches used neural networks for the individual classification of the instances. For example, Hansen et al.~\cite{T1-Hansen:2019} learned domain-specific word embeddings and syntactic dependencies and applied an LSTM classifier.

\noindent Using some external knowledge paid off ---they pre-trained the network with previous Trump and Clinton debates, supervised weakly with the ClaimBuster system.
Some efforts were carried out in order to consider context. Favano et al.~\cite{T1-T2-Favano:2019} trained a feed-forward neural network, including the two previous sentences as context. 
Whereas many approaches opted for embedding representations, feature engineering was also popular~\cite{T1-Gasior:2019}. 

\vspace{-0.25cm}
\paragraph{Task~2$_{2019}$.} \emph{Given a claim and a set of Web pages potentially relevant with respect to the claim, identify which of the pages (and passages thereof) are useful for assisting a human in fact-checking the claim. Finally, determine the factuality of the claim.}

The systems for evidence passage identification followed two approaches. BERT was trained and used to predict whether an input passage is useful to fact-check a claim~\cite{T1-T2-Favano:2019}. Other participating systems used classifiers (e.g., SVM) with a variety of features including similarity between the claim and a passage, bag of words, and named entities~\cite{T2Haouari:2019}. As for predicting claim veracity, the most effective approach used a textual entailment model. The input was represented using word embeddings and external data was also used in training~\cite{T2-Ghanem:2019}.

In the 2020 edition, Task~1$_{2019}$ becomes Task~5, and Task~1 is a reformulation based on tweets (cf.\ Section~\ref{sub:task1}). See~\cite{clef-checkthat-T1:2019} for further details. 
Task~2$_{2019}$ becomes Tasks~3 and~4 (cf.\ Sections~\ref{sub:task3} and ~\ref{sub:task4}). See~\cite{clef-checkthat-T2:2019} for further details. 

\vspace{-0.25cm}
\subsection{\ct 2018}
The 2018 edition featured two tasks~\cite{clef2018checkthat}:

\vspace{-0.25cm}
\paragraph{Task~1$_{2018}$} was identical to Task~1$_{2019}$.

The most successful approaches used either a multilayer perceptron or an SVM. Zuo et al.~\cite{T1-Zuo:2018} enriched the dataset by producing \textit{pseudo-speeches} as a concatenation of all interventions by a debater. They used averaged word embeddings and bag-of-words as representations. Hansen et al.~\cite{T1-Hansen:2018} represented the entries with embeddings, part of speech tags, and syntactic dependencies. They used a GRU neural network with attention. See~\cite{clef-checkthat-T1:2018} for further details. 

\vspace{-0.25cm}
\paragraph{Task~2$_{2018}$.} \emph{Given a check-worthy claim in the form of a (transcribed) sentence, determine whether the claim is likely to be true, half-true, or false.}

The best way to address this task was to retrieve relevant information from the Web, followed by a comparison to the claim in order to assess its factuality.\footnote{While this year a similar procedure had to be carried out, we decompose it into three tasks (cf.\ Section~\ref{sec:tasks}).}
After retrieving such \textit{evidence}, it is fed into the supervised model, together with the claim in order to assess its veracity. In the case of~\cite{T1-Hansen:2018}, they fed the claim and the most similar Web-retrieved text to convolutional neural networks and SVMs. Meanwhile, Ghanem et al.~\cite{T1-Ghanem:2018} computed features, such as the similarity between the claim and the Web text, and the Alexa rank for the website.
See~\cite{clef-checkthat-T2:2018} for further details.

\section{Related Work}
\label{sec:related}

There has been work on checking the factuality/credibility of a claim, of a news article, or of an information source~\cite{ba2016vera,R17-1046,ma2016detecting,mukherjee2015leveraging,popat2016credibility,zubiaga2016analysing}. Claims can come from different sources, but special attention has been given to those from social media~\cite{gupta2014tweetcred,mitra2015credbank,shu2017fake,zhao2015enquiring}. 
Check worthiness estimation is still a fairly-new problem especially in the context of social media ~\cite{RANLP2017:debates,Hassan:15,Hassan2016ComparingAF,hassan2017claimbuster}.

\ct further shares some aspects with other initiatives that have been run with high success in the past, e.g.,~stance detection (Fake News\footnote{Official Challenge website: \url{http://www.fakenewschallenge.org/}}), semantic textual similarity (STS at SemEval\footnote{STS task at the SemEval 2017 edition: \url{http://alt.qcri.org/semeval2017/task1/}}), and community question answering (cQA at SemEval\footnote{cQA task at the SemEval 2017 edition: \url{http://alt.qcri.org/semeval2017/task3/}}). 




\section{Conclusion}
\label{sec:final}

We have presented the 2020 edition of the \ct Lab, which features tasks that span the full verification pipeline: from spotting check-worthy claims to checking whether they have been fact-checked elsewhere already, to retrieving useful passages within relevant pages, to finally making a prediction about the factuality of a claim. To the best of our knowledge, this is the first shared task that addresses all steps of the fact-checking process.
Moreover, unlike previous editions of the \ct Lab, our main focus here is on social media, which are the center of ``fake news'' and disinformation. We further feature a more realistic information retrieval scenario with pooling for evaluation, as done at IR venues such as TREC. Last but not least, in-line with the general mission of CLEF, we promote multi-linguality by offering our tasks in different languages.

We hope that these tasks and the associated datasets will serve the mission of the \ct initiative, which is to foster the development of datasets, tools and technology that would enable the automatic verification of claims and will supporting human fact-checkers in their fight against ``fake news'' and disinformation.

\section*{Acknowledgments}
The work of Tamer Elsayed and Maram Hasanain was made possible by NPRP grant\# NPRP 11S-1204-170060 from the Qatar National Research Fund (a member of Qatar Foundation). The work of Reem Suwaileh was supported by GSRA grant\# GSRA5-1-0527-18082 from the Qatar National Research Fund and the work of Fatima Haouari was supported by GSRA grant\# GSRA6-1-0611-19074 from the Qatar National Research Fund. The statements made herein are solely the responsibility of the authors. This research is also part of the Tanbih project, developed by the Qatar Computing Research Institute, HBKU and MIT-CSAIL,
which aims to limit the effect of ``fake news'', propaganda, and media bias.

\bibliography{sigproc,clef19_checkthat,clef18_checkthat} 
\bibliographystyle{splncs04}
\end{document}